\documentclass[conference]{IEEEtran}
\IEEEoverridecommandlockouts

\usepackage{cite}
\usepackage{amsmath,amssymb,amsfonts}
\usepackage{graphicx}
\usepackage{textcomp}
\usepackage{xcolor}

\usepackage{graphicx}
\usepackage{amsmath, amssymb, amsthm}
\usepackage{amsfonts}
\usepackage{url}
\usepackage{algpseudocode}
\usepackage{algorithm}
\usepackage{booktabs}

\usepackage[margins]{trackchanges}

\addeditor{LV}

\begin{document}

\title{DODO: Causal Structure Learning with\\Budgeted Interventions}



\author{
    \IEEEauthorblockN{
        Matteo Gregorini\IEEEauthorrefmark{1}\IEEEauthorrefmark{2}, Chiara Boldrini\IEEEauthorrefmark{1}, Lorenzo Valerio\IEEEauthorrefmark{1}
        \thanks{C.~Boldrini and L.~Valerio contributed equally to this work.}
    }
    \IEEEauthorblockA{\IEEEauthorrefmark{1} IIT-CNR, via Moruzzi 1, Pisa, Italy}
    \IEEEauthorblockA{\IEEEauthorrefmark{2} University of Pisa, Pisa, Italy}
    \IEEEauthorblockN{\textit{\{name.surname\}@iit.cnr.it}}
}

\maketitle

\begin{abstract}

Artificial Intelligence has achieved remarkable advancements in recent years, yet much of its progress relies on identifying increasingly complex correlations. Enabling causality awareness in AI has the potential to enhance its performance by enabling a deeper understanding of the underlying mechanisms of the environment. In this paper, we introduce DODO, an algorithm defining how an Agent can autonomously learn the causal structure of its environment through repeated interventions. We assume a scenario where an Agent interacts with a world governed by a \emph{causal} Directed Acyclic Graph (DAG), which dictates the system's dynamics but remains hidden from the Agent. The Agent’s task is to accurately infer the causal DAG, even in the presence of noise. To achieve this, the Agent performs interventions,  leveraging causal inference techniques to analyze the statistical significance of observed changes. Results show better performance for DODO, compared to observational approaches, in all but the most limited resource conditions. 
DODO is often able to reconstruct with as low as zero errors the structure of the causal graph. In the most challenging configuration, DODO outperforms the best baseline by +0.25 F1 points.




\end{abstract}

\begin{IEEEkeywords}
Causal Theory, Causal Graph Discovery, Directed Acyclic Graphs
\end{IEEEkeywords}

\section{Introduction}\label{sec:intro}

Over the past decade, Artificial Intelligence has achieved remarkable advancements across diverse tasks, largely through powerful yet opaque ``black‑box'' deep learning models, raising concerns about their interpretability in critical applications \cite{LeCun_2015,Rudin_2019}. In addition, deep learning excels at finding correlations in static datasets but lacks the ability to capture cause-and-effect, imagination, reasoning, and planning, which limits AI’s capacity to generalize and transfer knowledge across environments~\cite{YoshuaBengioRevered}.
Causality theory offers a principled alternative: focusing on discerning true cause–effect relationships rather than mere correlations \cite{Pearl_2009, hernan2023causal, Guo_2020, zeng2023survey, deng2023causal, scholkopf2022causality, zhu2020causal, dasgupta2019causal}. This shift is crucial, as reliance on associations alone can easily lead to spurious or unsafe conclusions \cite{Spirtes_2000}.
Embedding causal reasoning into AI systems not only aids interpretability but also aligns model behavior with real‑world interventions \cite{Guo_2020}. As many researchers argue \cite{chauhan2025a, 10.1145/3665494,genomic, scholkopf2022causality, causality_engineering}, the next step for AI is to learn how to learn cause-and-effect relations.

Causal relationships are commonly formalized as Directed Acyclic Graphs (DAGs), where variables are connected by directed edges representing causal influence \cite{Shanmugam2015SmallInt}. For example, in healthcare, a DAG might represent how smoking leads to lung disease, which in turn increases the probability of hospitalization; here, the directed edges explicitly encode the causal pathways rather than mere correlations. In a smart home scenario, a DAG could represent how occupancy detection triggers heating or lighting systems, which then affects energy consumption.
By structuring knowledge within causal graphs, AI systems gain the ability to generate inherently explainable decisions \cite{Madumal_2020}. Potential applications for causal learning include industrial \cite{zhu2022causaldyna}, video games \cite{wang2023voyager}, health care \cite{applications_health} and recommendation systems \cite{nie2023knowledge}. The causality-aware industrial AI can be used for automated and smarter learning of strategies of robotic interactions with the environment. Video games provide an excellent testing and application ground based on the fact that they are highly controllable environments, widely available and diverse, and provide many different learning scenarios. 
Healthcare is a promising~\cite{pmlr-v119-zhang20a, mental_health_interventions, applications_health} but highly complex application area: ethical issues and data management problems (especially regarding patient privacy) alone make it the most difficult area to manage. Nonetheless, the literature has devoted considerable attention to interventions in this field.

A central challenge in causal discovery is the inference of causal structures from data. Approaches relying solely on observational data are inherently limited, as correlations alone cannot resolve equivalence classes of causal graphs~\cite{hernan2023causal}. Interventional data, obtained by actively perturbing the system, do significantly reduce ambiguity~\cite{Pearl_2009}, but practical algorithms for leveraging repeated interventions remain scarce. Existing techniques either require a large number of interventions, incur high computational complexity, or assume prior knowledge that may not be available in realistic settings.

To address these limitations, we introduce DODO, a novel approach for causal structure learning through repeated interventions. DODO iteratively selects and applies interventions, updating its estimate of the underlying causal graph in light of the new evidence. Unlike purely score-based or constraint-based methods, DODO exploits a lightweight heuristic to efficiently guide the intervention process, balancing exploration of uncertain edges with exploitation of already established structures. This results in improved accuracy with fewer interventions, making DODO suitable for resource-constrained environments where interventions are costly or limited.

Our main contributions are the following:
\begin{itemize}
    \item We introduce DODO, a heuristic framework for causal structure learning that integrates repeated interventions with observational analysis of linear, noisy systems.
    \item We evaluate DODO on synthetic datasets, demonstrating that it outperforms observational approaches and is able to uncover the underlying graph causal structure with excellent accuracy. In the most challenging configuration considered, DODO clearly dominates, reaching near-perfect F1, while baseline NOTEARS plateaus at ~0.7 and baseline PC stagnates at ~0.3.
    
\end{itemize}

The paper is organized as follows. In Section~\ref{sec:related_work}, we provide a brief overview of the state of the art. Section~\ref{sec:system_modelling} presents our system model. Section~\ref{sec:DODO} introduces our new algorithm, DODO. The settings of our evaluation framework are presented in Section~\ref{sec:experiments_settings}. Lastly, Section~\ref{sec:result} discusses the performance of the DODO algorithm against the chosen baselines.


\section{Background and Related Work}
\label{sec:related_work}

Causal theory leverages Structural Causal Models (SCMs) and Directed Acyclic Graphs (DAGs) to formalize cause–effect relationships through structural equations that partition variables into exogenous (external) and endogenous (derived) types, thus enabling reasoning about interventions via Pearl’s do‑operator~\cite{Pearl_2009}. Unlike traditional statistical models that rely solely on passive observation, causal theory allows active manipulation, symbolized by expressions like \(\mathrm{do}(A=a)\), and supports answering counterfactual “what if” questions by altering assumed causal mechanisms. Core DAG structures—chains (A→B→C), forks (A←B→C), and colliders (A→B←C)—govern how associations appear or vanish, revealing which dependencies are genuine causal links and which arise from confounding or conditioning effects. 

\subsection{Learning causality via observational data}

Without interventional data, structure learning relies on statistical regularities of the joint distribution and typically returns a Markov equivalence class (i.e., the set of all DAGs that encode exactly the same conditional independence relations), rather than a unique DAG~\cite{Spirtes_2000}. 

\textit{Constraint-based} procedures (e.g., \text{PC}\cite{SpirtesGlymour1991PC}, \text{FCI}\cite{spirtes2001anytime}/\text{RFCI}\cite{Colombo2012RFCI}) infer graphs by testing conditional independencies and orienting edges to satisfy the discovered constraints, yielding CPDAGs (Completed Partially Directed Acyclic Graph), representing the Markov equivalence class, with directed edges for those present in every DAG and undirected ones for those with different orientations; or PAGs (Partial Ancestral Graph), representing a Markov equivalence class of DAGs incorporating uncertainty caused by unobserved nodes~\cite{Spirtes_2000,Zhang2008FCI,Colombo2012RFCI}.

\textit{Score-based} approaches frame discovery as an optimization over equivalence classes; GES\cite{Chickering_2002}/FGES~\cite{Ramsey2017FGES} maximize a decomposable, complexity-penalized score, while continuous formulations such as NOTEARS\cite{Zheng2018NOTEARS} enforce acyclicity via a differentiable constraint for gradient-based optimization. Functional causal models exploit asymmetries in mechanisms and noise to orient edges from purely observational data: LiNGAM~\cite{Shimizu2006LiNGAM} achieves full identifiability, meaning that the causal discovery determines one single causal model instead of an equivalent class, under non-Gaussian disturbances; additive-noise models extend this idea to nonlinear relations \cite{Shimizu2006LiNGAM,Hoyer2009ANM,Peters2014ANM}. Information-theoretic and kernel-based techniques strengthen testing/orientation by using dependence measures, for example kernel conditional-independence tests (KCI) and HSIC-based tests for nonlinear and continuous settings \cite{Zhang2011KCI,Gretton2005HSIC,Gretton2007HSIC}.

DODO advances beyond purely observational approaches by introducing a heuristic for leveraging interventions. As demonstrated in Section~\ref{sec:result}, this substantially enhances the accuracy of causal structure learning.

\subsection{Learning causality through interventions}

Interventional causal discovery uses experimental manipulations to break observational equivalence and refine identifiability. While observational data often leave multiple DAGs indistinguishable within a Markov equivalence class, interventions can orient edges that would otherwise remain ambiguous. Formally, an \emph{interventional Markov equivalence class} (I-MEC) collects all DAGs consistent with both observational and interventional distributions~\cite{Hauser2012GIES}. 

Classical results in experimental design establish upper bounds on the number of interventions required for full identifiability. Under the assumption of \emph{causal sufficiency}---i.e., that there are no hidden confounders among the measured variables---and \emph{perfect single-variable interventions}, where one can deterministically set the value of a variable while leaving all other mechanisms unchanged, at most $N-1$ experiments suffice to recover a causal graph over $N$ variables. In our case, the Agent does not operate under the assumption of observing a Directed Acyclic Graph necessarily, rendering the $N-1$ limit not bounding for our study. Allowing \emph{multi-variable interventions} (simultaneous manipulation of several nodes) reduces this worst-case requirement exponentially to $\lceil \log_2 N\rceil + 1$~\cite{Eberhardt2006Nminus1,Eberhardt2012LogN}.
These results highlight how interventions can dramatically shrink the search space compared to purely observational discovery.  

Algorithmically, score-based methods such as GIES~\cite{Hauser2012GIES} extend greedy equivalence search to interventional settings, combining observational and experimental data to estimate I-MECs. Active-learning strategies go further by selecting interventions adaptively: early approaches use minimax or entropy-based criteria to reduce uncertainty~\cite{HeGeng2008JMLR}, while later work designs interventions that optimally shrink undirected cliques or maximize the number of orientable edges~\cite{Hyttinen2013JMLR,Hauser2014IJAR}. When large-scale manipulations are infeasible, ``small'' interventions that target only a few variables at a time can still guarantee near-minimal experiments for full orientation~\cite{Shanmugam2015SmallInt}.  

Beyond explicit intervention design, the \emph{Joint Causal Inference} (JCI) framework integrates observational data with diverse interventional contexts, including cases with unknown or overlapping targets, within a unified formalism. This enables algorithm-agnostic discovery without requiring precise intervention labels~\cite{Mooij2020JCI}.  

In machine learning, continuous optimization approaches have been extended to incorporate interventions. For instance, \textsc{DCDI} integrates interventional information into differentiable score-based discovery, scaling to high-dimensional graphs and accommodating perfect, imperfect, or even unknown-target interventions~\cite{Brouillard2020DCDI}. Related approaches leverage invariance principles to handle \emph{soft interventions}, where only the data-generating mechanism changes (e.g., regression function) but the distribution of exogenous noise may not be fixed, and characterize the resulting equivalence classes~\cite{Jaber2020Soft}.  

Finally, active experimentation has been reframed as a sequential decision problem. \emph{Causal bandits} extend multi-armed bandit algorithms by exploiting partial knowledge of a causal graph, thereby achieving lower regret through targeted interventions~\cite{Lattimore2016CausalBandits}. Neural active-intervention methods push this further, combining structure learning with experiment selection in an online setting~\cite{Scherrer2021ActiveInt}.  

\subsection{Graph Sizes and Structures in Causal Discovery}
\label{sec:relwork_graphs}

An extensive review of the causal discovery literature reveals a broad spectrum of graph sizes and structures. Most studies focus on graphs with 10--20 nodes or up to 50--100 nodes~\cite{Glymour2019Review, Sachs2005Science, LauritzenSpiegelhalter1988, Cooper}, though applications in domains such as genomics and social networks have extended analyses to graphs with hundreds or even thousands of variables~\cite{Bansal_2007, Aicher_2014}. 

The underlying formalism of causal graphs is that of directed acyclic graphs (DAGs), which forbid directed cycles and impose a strict partial ordering on nodes. As a result, the maximum number of possible edges is bounded by $\binom{N}{2}$~\cite{Robinson_1973, Koller_Friedman_2009}, with $N$ being the number of nodes.  
The density of DAGs---and hence the expected number of edges---depends on edge probabilities, graph construction models, and domain-specific constraints. Sparse DAGs, often modeled via adjusted Erd\H{o}s--R\'enyi random graphs, scale roughly linearly with the number of nodes, while dense DAGs approach the combinatorial upper bound~\cite{Meka_Pitassi_2009, Dagum_Luby_1992}. Empirical studies confirm that real-world causal structures are typically sparse, a property consistently observed in biological systems and temporal social networks~\cite{Bansal_2007, Aicher_2014}.  

In artificial intelligence and machine learning, both the scalability and accuracy of causal discovery algorithms are highly sensitive to graph size and density. Chickering~\cite{Chickering_2002} showed how the expressiveness of Bayesian networks comes at the cost of computational hardness, while Koller and Friedman~\cite{Koller_Friedman_2009} highlighted the intractability of inference in dense structures. Standardized benchmarks, such as the \textsc{Alarm} and \textsc{Child} networks, have become cornerstones for empirical evaluation and comparison of algorithms~\cite{Chickering_2002}.  

Despite continuous algorithmic progress, learning causal structures on large-scale DAGs remains a formidable challenge. Applications in genomics, for example, often involve thousands of variables, far beyond the typical settings of benchmark datasets~\cite{Bansal_2007}. These insights guided our design choices: we target small to moderately sized DAGs---where scalability is tractable and empirical validation feasible---while acknowledging that handling large, dense graphs is a critical open problem for future work.

\begin{table}[!t]
\centering
\caption{Mathematical Notation}
\label{tab:notation}
\begin{tabular}{ll}
\toprule
\textbf{Notation} & \textbf{Description} \\
\midrule
$N$ & Number of nodes\\
$\mathcal{V}$ & Set of nodes, $\mathcal{V} = \{v_1, v_2, \dots, v_N\}$\\
$\mathcal{G}_{ER}$ & Directed Erdos Renyi Graph\\ 
$\mathcal{G}_{DAG}$ & Directed Acyclic Graph (true causal structure) \\ 
$\mathcal{E}_{DAG}$ & Set of direct causal edges\\
$X_v$ & Random variable representing the value of node $v$\\ 
$X_v^{(k)}$ & Observed value of node $v$ at epoch $k$ (obs. phase)\\ 
$X_v^{int(k)}$ & Intervened value of node $v$ at epoch $k$ (interv. phase)\\ 
$B$ & Budget, i.e., total number of epochs at the Agent disposal \\ 
$K$ & Number of epochs in each phase \\ 
$\hat{\mu}_v^{obs}$ & Expected value of node $v$ from observational data \\ 
$\hat{\mu}_v^{int}$ & Expected value of node $v$ under intervention\\ 
$\mathcal{E}_{cand}$ & Set of candidate causal edges after intervention \\ 
$p_{crit}$ & Significance level threshold for Welch's t-test \\ 
$p_{val}(\cdot,\cdot)$ & P-value from Welch’s t-test comparing distributions \\ 
$\mathcal{S}_{uv}$ & Set of nodes conditioning $u$, excluding $v$ \\ 
$\hat{\rho}_{X_u,X_v | \mathcal{S}_{uv}}$ & Empirical partial correlation of nodes $u,v$ given $\mathcal{S}_{uv}$ \\ 
$\mathcal{E}_{Agent}$ & Final inferred set of direct causal edges after pruning \\ 
$\mathcal{G}_{Agent}$ & Final inferred directed acyclic causal graph\\ 
${w_uv}$ & Causal strength from node $u$ to node $v$ \\ 
$\text{Pa}(v)$ & Set of parent nodes of node \(v\)\\
$\hat{\text{Pa}}(v)$ & Agent's belief of parent nodes of node \(v\)\\
$\mathcal{D}_{obs}$ & Observational dataset used for parameter estimation \\ 
$\mathcal{A}_{DAG}$ & Ground truth binary adjency matrix \\ 
$\mathcal{A}_{Agent}$ & Inferred binary adjency matrix \\ 
$\mathcal \lambda_v^2$ & Std. Dev. of Gaussian Noise Distribution \\ 
$\mathcal \mu_v$ & Mean of the Normal Distribution for node v \\ 
$\mathcal \sigma_v^2$ & Std. Dev. of Normal distribution for node v \\ 
$\mathcal \tau$ & Value of Intervention \\ 

\bottomrule
\end{tabular}
\end{table}

\section{System model}\label{sec:system_modelling}

We consider a World, and an Agent tasked with uncovering the causal structure governing it.
We assume that the World evolves according to the rules defined in Section~\ref{sec:world_model}. Outside of it, the Agent can observe the state of the World and interact with it through a set of predefined actions, as described in Section~\ref{sec:agent_model}.

\subsection{World model}
\label{sec:world_model}



Let the World be represented by a set of $N$ nodes, denoted by $\mathcal{V} = {v_1, v_2, \dots, v_N}$. The relationships among these nodes are captured by an acyclic, directed graph, denoted by $\mathcal{G}_{DAG}=(\mathcal{V},\mathcal{E})$, where $\mathcal{E}$ denotes the set of the edges connecting nodes in $\mathcal{V}$. Each node $v \in \mathcal{V}$ corresponds to a random variable $X_v \in \mathbb{R}$, whose value is defined through a Structural Causal Model (SCM)~\cite{Pearl_2009}.
An SCM defines each variable as a function of its parent variables in the graph, plus an independent noise term, i.e., 
\begin{equation} \label{eq:sem}
X_v = f_v\big(\{X_u : u \in \text{Pa}(v)\}, \, \epsilon_v\big),
\end{equation}
where \(\text{Pa}(v)\) denotes the set of parent nodes of \(v\) in \(\mathcal{G}_{DAG}\), \(f_v(\cdot)\) is a deterministic function, and \(\epsilon_v\) is an exogenous noise random variable ($\epsilon_v$ are assumed to be mutually independent across nodes $v$). This formulation provides the causal semantics of the model: performing an intervention on a variable \(X_v\) corresponds to replacing its structural equation in Eq.~\eqref{eq:sem} with an externally imposed value \(X_v = x\). In Pearl’s framework~\cite{Pearl_2009}, an intervention is formalized by the do-operator, \(\mathrm{do}(X_v = x)\), which alters the generative process for \(X_v\) while leaving the equations for all other variables unchanged.

In our system, a node without incoming edges—referred to as a source or exogenous node—is assigned its value based on an initial random draw from a normal distribution. Specifically, for any source node $v$, where the set of parent nodes $\text{Pa}(v)$ is empty, we have:
\begin{equation}
X_v = \gamma_v + \epsilon_v
\end{equation}
where $\gamma_v \sim \mathcal{N}(\mu_v, \sigma_v^2)$ and $\epsilon_v \sim \mathcal{N}(0, \lambda_v^2)$ represent node-specific Gaussian random variables. Parameters $\mu_v$ and $\sigma_v^2$ are initialized individually for each source node. 
%
Nodes with incoming edges, termed non-source or endogenous nodes, derive their values through linear combinations of their parent nodes' values, combined with the standard Gaussian noise $\epsilon_v$. Thus, for any node $v$ with $\text{Pa}(v) \neq \emptyset$, we have:
\begin{equation}
X_v = \sum_{u \in \text{Pa}(v)} w_{uv} X_u + \epsilon_v
\end{equation}
Here, coefficient $w_{uv} \in \mathbb{R}$ represents causal influence from node $u$ to node $v$.
%
%
The explicitly acyclic nature of $\mathcal{G}_{DAG}$ guarantees a unique solution for the system of equations describing all $X_v$ values. Consequently, the defined SCM captures instantaneous equilibrium conditions. 

We adopt a Normal distribution for the variables as it is standard in linear–Gaussian structural equation models, where identifiability and estimation are well established~\cite{PetersBuhlmann2014,ParkKim2020,KalischBuhlmann2007,GeigerHeckerman1994,PetersJanzingSchoelkopf2017,Shimizu2006LiNGAM,Koller_Friedman_2009}. Normality also offers a reasonable approximation of aggregated measurement perturbations and latent fluctuations in empirical causal graphs~\cite{Spirtes_2000,Koller_Friedman_2009,lauritzen1996graphical,carroll2006measurement,feller1971introduction,bollen1989structural}. Moreover, given the combinatorial burden of graph topologies and random seeds, fixing the noise family avoids further complexity. We thus take the Gaussian case as a natural baseline, to be later relaxed to alternative distributions. We restrict ourselves to linear relationships because linear SCMs form a widely studied core model class in causal discovery, with well-established algorithms and identifiability results~\cite{KalischBuhlmann2007,PetersBuhlmann2014,ParkMoonParkJeon2021,PetersJanzingSchoelkopf2017}. This restriction also helps contain the combinatorial complexity of testing across multiple graph scenarios. Finally, as with distributional assumptions, we view linearity as a provisional baseline, with future work extending to richer (e.g., nonlinear) causal mechanisms once baseline behavior is established.


\subsection{Agent model}
\label{sec:agent_model}

The Agent observes and interacts with the World described in Section~\ref{sec:world_model}. Its goal is to reconstruct the SCM governing the system through repeated observations and interactions. An \emph{epoch} corresponds to one discrete step in which the Agent observes the state of the world (at rest or after an intervention). 
At each epoch, the Agent can access the states of all variables but has no prior knowledge of the underlying causal dependencies. The structure of causal connections is therefore hidden, and the system initially appears as an undifferentiated collection of observables. The environment’s dynamics involve no propagation delay: causal influences are realized instantaneously, and their effects are fully expressed at each iteration. As a result, the Agent perceives a synchronous unfolding of the system state across all nodes, without temporal offsets between cause and effect that could facilitate/hinder causal discovery. 

\section{The DODO algorithm}
\label{sec:DODO}


This paper proposes DODO, a causal discovery algorithm that drives the actions of the Agent to uncover the SCM governing the system. 
The DODO algorithm is divided into 4 phases: i) observation, ii) intervention, iii) detection and iv) pruning. In the first two phases, the algorithm observes and interacts with the environment, storing the observed values for each node in the graph. Secondly, the Agent compares the values at rest and under intervention, determining which changes brought persistent alterations to other nodes. Lastly, the Agent prunes all indirect causal connections by comparing the partial correlations. For the convenience of the reader, the notation used in this paper is summarised in Table~\ref{tab:notation}.

\subsection{Observation Phase}\label{sec:observation_dodo}

The goal of the \emph{observation phase} is to establish a baseline measurement of the system ``at rest’’, which can be later used by the Agent to assess the effect of an intervention. To this aim, for the first \(k = 1, \dots, K\) epochs, the Agent records the observed values exhibited by every node \(v\).
At the end of the phase, for each node $v$, the Agent computes the average value observed: 
\begin{equation}
\hat{\mu}_v^{obs} = \frac{1}{K}\sum_{k=1}^K X_v^{(k)}, \forall v\in\mathcal{V}.
\end{equation}

\subsection{Intervention Phase}\label{sec:interventions_dodo}

In the subsequent \emph{intervention phase}, the Agent performs a set of consecutive interventions, i.e., it actively perturbs each node to evaluate the resulting causal effects. Formally, for each node \(v \in \mathcal{V}\), the Agent intervenes by assigning node \(v\) a fixed value, overriding its naturally occurring value. In this paper, we set the intervention value as: 
\begin{equation}
    \tau =2 \max_{v = 1 \ldots N}|\hat{\mu}_v^{obs}|,
\end{equation}
In other words, we assign a value that is higher than the expected values of all variables when performing the intervention. This ensures that the intervention is clearly distinguishable from the natural values of the variables. As future work, we plan to explore alternative types of interventions.
Each such intervention is independently repeated for \(K\) epochs, yielding the intervened values \(X_v^{int(k)}\). 



\subsection{Causal Links Detection}\label{sec:causal_links_detection}
By comparing the values observed for the system ``at rest'' and those after intervention, the Agent can now make informed guesses about the existence of causal links in the system DAG.
%
A potential causal relationship from an intervened node $u^*$ to another node $v$ ($u^* \rightarrow v$) is posited if the distribution of $X_v$ under intervention on $u^*$ significantly differs from its observational distribution. This is assessed by using a statistical test, in our case the two-sample t-test on the means of the two variables $X_v$ and $X_{u*}$.

Let $\mathcal{E}_{cand}$ be the set of candidate directed edges. An edge $(u^*, v)$ is added to $\mathcal{E}_{cand}$ if the p-value $p_{val}$ of the test is smaller than a predefined significance level $p_{crit}$ (e.g., we use $0.001$ in this paper):
\begin{equation}
    p_{val} < p_{crit}.
\end{equation}
The set $\mathcal{E}_{cand}$ may include both direct and indirect causal links, depending on the graph topology (and in some cases, it may contain no indirect links at all). For example, in a chain $A \to B \to C$, single interventions on either $A$ or $B$ affect $C$, leading to both $A \to C$ and $B \to C$ being added to $\mathcal{E}_{cand}$, even though only the direct edge $B \to C$ should be retained.

\subsection{Indirect Causal Connections Pruning}\label{sec:pruning}

In this phase no more interventions or observations are performed. The Agent entirely relies on the data acquired in the previous phases. 
At the end of the previous phase, the Agent built a candidate causal graph $\hat{\mathcal{G}}_{DAG} = (\mathcal{N}, \mathcal{E}_{cand})$. We denote with $\hat{\text{Pa}}(u)$ the set of parent nodes to node $u$ in $\hat{\mathcal{G}}_{DAG}$, i.e., the nodes that may causally affect $u$.
To discern direct causal relationships from indirect ones mediated by other nodes, the Agent focuses on nodes $u$ that have more than one candidate parent, i.e., $u$ such that ${|\hat{\text{Pa}}(v)|} > 1$. For each candidate edge $(u, v) \in \mathcal{E}_{\text{cand}}$, the Agent computes the \textit{partial correlation} between $X_u$ and $X_v$ while controlling for all other possible causes, i.e., all other nodes in $\hat{\text{Pa}}(u)$ that are not $v$.
The Agent performs a partial correlation to verify whether the node that is being deemed as a possible direct cause is contributing unique information to the node that is being studied as the effect.
We denote the conditioning set as $\mathcal{S}_{uv} = \hat{\text{Pa}}(u) \setminus \{v\}$. The sample partial correlation is $\hat{\rho}_{X_u,X_v | \mathcal{S}_{uv}}$.
A hypothesis test is performed:
\begin{itemize}
    \item $H_0$: $\rho_{X_u, X_v | \mathcal{S}_{uv}} = 0$ (i.e., $X_u$ and $X_v$ are conditionally independent given $\mathcal{S}_{uv}$, suggesting no direct causal edge $u \to v$ if $\mathcal{S}_{uv}$ d-separates them or contains all other true parents).
    \item $H_1$: $\rho_{X_u, X_v | \mathcal{S}_{uv}} \neq 0$.
\end{itemize}
Let $p_{val}(u,v)$ be the p-value obtained from this test. If $p_{val}(u,v) > p_{crit}$, where $p_{crit}$ is the aforementioned pre-set acceptable error threshold (significance level), the Agent fails to reject $H_0$. In this scenario, the relationship $(u,v)$ is considered indirect, and the direct edge $(u,v)$ is deemed non-existent.
The set of inferred direct causal edges is $\mathcal{E}_{Agent} = \{ (u,v) \in \mathcal{E}_{cand} \mid p_{val}(u,v) \le p_{crit} \}$. The Agent constructs the final inferred directed acyclic graph $\mathcal{G}_{Agent} = (\mathcal{V}, \mathcal{E}_{Agent})$.



\subsection{Budget and Epochs}
The time budget of the Agent is the total amount of epochs that the Agent can use to interact with the World. The Agent allocates its budget across the different phases. Each interaction with the environment (whether an observation or an intervention) must be repeated $K$ times to obtain sufficiently many samples for reliable t-tests. Accordingly, during the observation phase, the Agent observes the World for $K$ epochs. Then, in the intervention phase, the Agent applies one intervention per node and again collects $K$ samples for each. In total, this results in $K(N+1) = B$, where $B$ is the available budget.
The number of epochs $K$ available for each distinct interaction follows as:
\begin{equation}
    K = \left\lfloor \frac{B_{total}}{N+1} \right\rfloor 
\end{equation}
It follows that $K$ is larger (leading to better results for the statistical tests) when $B$ is larger or $N$ is smaller.

\section{Experimental settings}
\label{sec:experiments_settings}

\subsection{Synthetic DAG Generation}
\label{sec:dag_generation}

As discussed in Section~\ref{sec:world_model}, the causal structure of the world to be uncovered by the Agent is represented by a SCM and its associated DAG. To explore different configurations, for our experiments we generate synthetic ground-truth DAGs \(\mathcal{G}_{DAG} = (\mathcal{V}, \mathcal{E}_{DAG})\), and we vary $N$ ($=|\mathcal{V}|$) in  $\{5,\,10,\,20\}$. We chose these sizes because of their wide usage in scientific literature \cite{Sachs2005Science, Mooij2016Benchmark, Peters2014ANM}. Each DAG is obtained from a directed Erdős--Rényi graph~\cite{barabasi2013network} where every possible ordered node pair \((u,v)\), \(u \neq v\), is included independently with probability \(p \in \{0.15,\,0.30,\,0.50\}\). Acyclicity is enforced by iteratively removing one edge per detected directed cycle until none remain, ensuring that \(\mathcal{G}_{DAG}\) admits a topological ordering \(\pi\). 
The weight $w_{uv}$ of each edge $(u, v)$ is drawn from a mixture of two uniform distributions; with equal probability, the weight is sampled either from $\mathcal{U}(-5, -1)$, or from $\mathcal{U}(1, 5)$. This choice allows us to test both causal and anticausal effects.




\subsection{Simulations settings}
\label{sec:simulation_settings}

For each independent node, the value of the mean for the Normal distributions was extracted uniformly at random in $[-50, 50]$, while the standard deviation was extracted uniformly at random in $[1,2]$. Constraining exogenous variables to a modest standard deviation renders the model more realistic, by reflecting the inherent stability of real-world foundational factors. This prevents the propagation of unrealistic volatility through the causal graph, yielding a more accurate representation of the data-generating process.

For each combination of node size \(N\in\{5,10,20\}\) and edge probability \(p\in\{0.15,0.30,0.50\}\), we generate 30 different random graph topologies using independent random seeds. We also systematically diversify our synthetic DAGs, mapping them to different SCMs, by introducing three distinct levels of Gaussian noise intensity to reflect varying degrees of observational uncertainty. The Gaussian noise was generated by drawing a value from a Normal distribution with mean $0$ and a standard deviation obtained by multiplying a noise coefficient (i.e., taken from the set $\{0.0001, 0.5, 1\}$) by a fixed standard deviation measure. Depending on the chosen noise coefficient, the resulting noise level may be lower, comparable, or higher than the standard deviation magnitude of the Normal distributions of the nodes.
To assess algorithmic performance under resource constraints, we enforce ten distinct budget levels for the maximum total budget, \(B \in \{100,200,\dots,1000\}\), spanning from low to relatively high budget regimes.  
Furthermore, for each simulation we employ 30 different seeds for random number generation governing the initialization of the independent node values, as well as the causal edges weights, to ensure replicability and variability across trials, leading to a total of 900 runs for each combination of $p$, noise intensity, $B$ and $N$. 
The results are reported as the average across these 900 runs, with a 95\% confidence interval.

\section{Performance Evaluation}

\subsection{Performance Metrics}

Let $\mathbf{A}_{DAG} \in \{0,1\}^{|\mathcal{V}|\times|\mathcal{V}|}$ denote the binary adjacency matrix of the ground-truth DAG $\mathcal{G}_{DAG} = (\mathcal{V}, \mathcal{E}_{DAG})$, where
\[
(\mathbf{A}_{DAG})_{ij} =
\begin{cases}
1, & \text{if } (v_i, v_j) \in \mathcal{E}_{DAG}, \\
0, & \text{otherwise}.
\end{cases}
\]
Similarly, let $\mathbf{A}_{Agent} \in \{0,1\}^{|\mathcal{V}|\times|\mathcal{V}|}$ denote the binary adjacency matrix of the inferred graph $\mathcal{G}_{Agent} = (\mathcal{V}, \mathcal{E}_{Agent})$. The performance of the causal discovery procedure is evaluated by comparing $\mathbf{A}_{Agent}$ to $\mathbf{A}_{DAG}$, with particular attention to how reconstruction accuracy varies as a function of the total number of available epochs $N$.
The F1 score and Structural Hamming Distance are computed. 
\paragraph{F1 score} The F1 score's is helpful in cases of class imbalance, such as the causal graphs under consideration, where the number of present causal links ($\mathbf{A}_{Agent}$) is significantly outnumbered by the potential non-existent links. It thus gives appropriate importance to the correct identification of these present links.

\paragraph{Structural Hamming Distance - SHD)} The number of edge additions or deletions required to transform $\mathbf{A}_{Agent}$ into $\mathbf{A}_{DAG}$. This corresponds to the sum of false positives (FP) and false negatives (FN) when considering only edge presence/absence.
\begin{equation}
    \text{SHD} = \text{FP} + \text{FN}
\end{equation}
The SDH provides an overall, unweighted measure of the structural difference between the inferred and true graphs, giving a comprehensive picture of the prediction's correctness.

The Structural Hamming Distance offers a global perspective on the structural accuracy, while the F1 score provides a more nuanced evaluation that prioritizes the successful recovery of existing causal connections.

\subsection{Comparative Baselines}\label{sec:obs_baselines}

To benchmark the Agent's performance, its causal discovery capabilities are compared against a diverse set of established and contemporary observational causal discovery algorithms. Each algorithm has been tested with a range of hyperparameters, and the ones leading to the highest F1 score configuration was used.  The suite of baselines includes:

\begin{figure}[!t]
\centering
\includegraphics[width=1\linewidth]{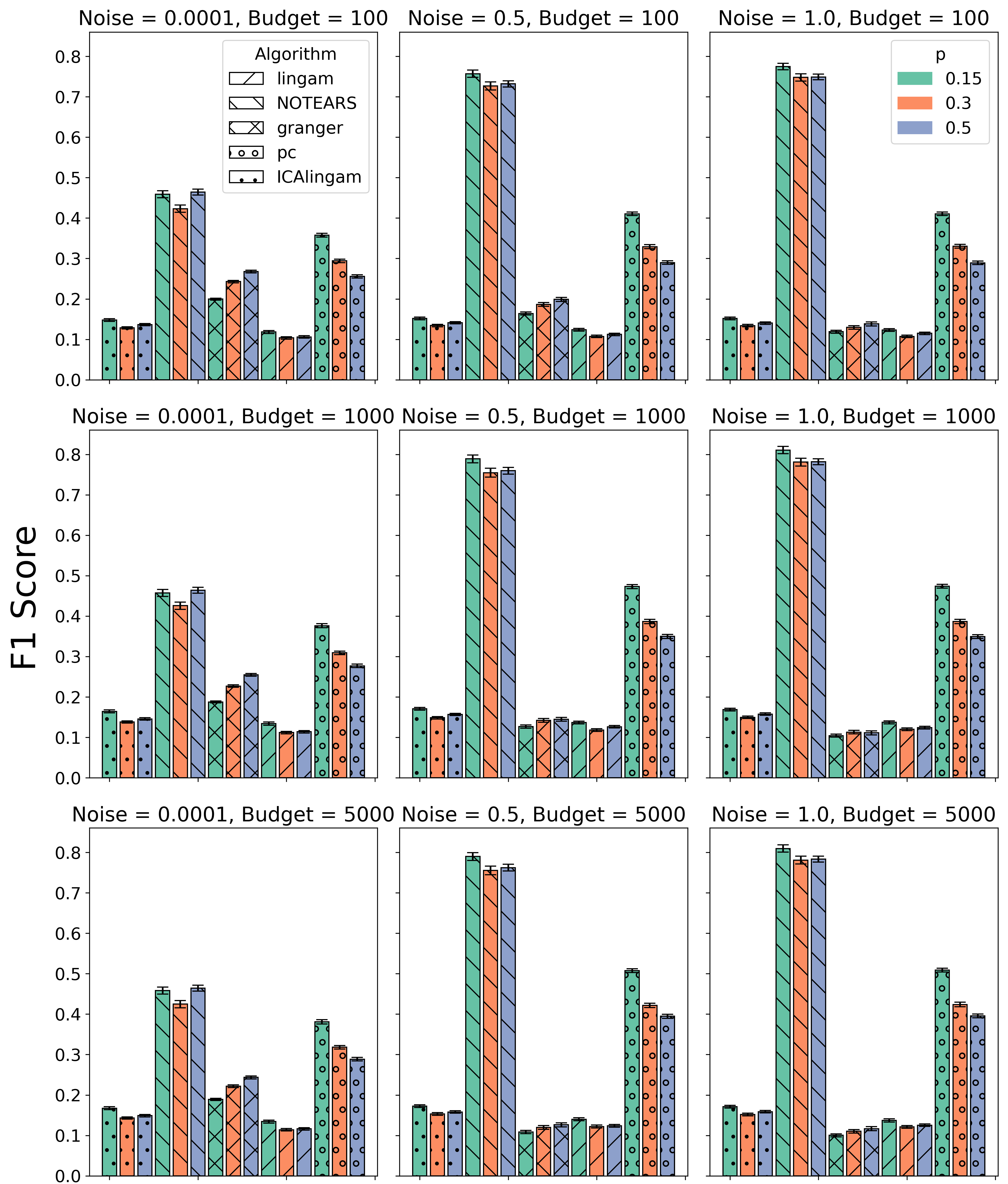}
\caption{Observational Algorithms F1 score comparison with 95\% confidence interval }
\label{fig:20_nodes_obs}
\end{figure}

\textbf{Granger Causality}\cite{Granger_1969}, a statistical hypothesis testing approach primarily for time-series data. It infers that a variable $X$ Granger-causes $Y$ if past values of $X$ significantly improve the prediction of $Y$. This involves comparing a restricted autoregressive model for $Y_t$ with an unrestricted one including lagged $X_t$:
\begin{equation}
    Y_t = \sum_{i=1}^p \alpha_i Y_{t-i} + \sum_{j=1}^q \beta_j X_{t-j} + \epsilon_t
\end{equation}
A statistical test (e.g., F-test) is performed on $H_0: \beta_1 = \dots = \beta_q = 0$. Three tiers of maximum lag were used to determine the best performance. 

The \textbf{Peter-Clark (PC) algorithm}\cite{SpirtesGlymour1991PC}, a constraint-based method. It starts with a fully connected undirected graph and prunes edges based on conditional independence tests. An edge between nodes $X_i$ and $X_j$ is removed if they are found to be conditionally independent given some subset $\mathbf{S} \subseteq \mathcal{V} \setminus \{X_i, X_j\}$:
\begin{equation}
    X_i \perp\!\!\!\perp X_j | \mathbf{S}
\end{equation}
Subsequently, it orients edges based on d-separation principles and collider detection. We use three conditional-independence tests: Fisher’s z-transform of partial correlations for Gaussian data, Pearson’s chi-square test for counts, and the likelihood-ratio G-test, approximating chi-square.

\textbf{LiNGAM (Linear Non-Gaussian Acyclic Model)}\cite{Shimizu2006LiNGAM}, which identifies causal structure under the assumption of linear relationships and non-Gaussian, independent error terms. The model is typically expressed as as $\mathbf{x} = (\mathbf{I} - \mathbf{B})^{-1} \mathbf{e}$, where $\mathbf{B}$ is a weighted adjacency matrix that can be permuted to be strictly lower triangular, and components of $\mathbf{e}$ are non-Gaussian and mutually independent. We employed two different statistical methods, each used to determine the independence between variables: the Pairwise-LiNGAM and the Kernel based one (the latter using the Hilbert-Schmidt Independence\cite{Gretton2005HSIC}).\par
\textbf{ICALiNGAM}\cite{Shimizu2006LiNGAM}, a variant of LiNGAM that integrates ICA more tightly into the estimation process. It seeks a permutation of the ICA components that best matches a causal ordering, improving robustness and accuracy. The model retains the same structural assumptions as LiNGAM but enhances estimation stability through iterative refinement of the ICA-based decomposition. We vary the parameter \texttt{max\_iter} in ICALiNGAM (set to \{$10^3, 10^4, 10^5$\}) to control the maximum number of iterations allowed for its FastICA subroutine.

\textbf{NOTEARS}\cite{Zheng2018NOTEARS}, which formulates causal discovery as a continuous constrained optimization problem. For linear models, it aims to find a weighted adjacency matrix $\mathbf{W}$ by minimizing a loss function (e.g., squared error) subject to an algebraic acyclicity constraint. We vary the hyperparameter \texttt{max\_iter} in NOTEARS (set to $\{5, 10, 20\}$) to control the maximum number of dual ascent steps during optimisation.

These algorithms provide a robust set of comparators, spanning different theoretical assumptions and methodological approaches to observational causal inference. 
We conducted a preliminary set of simulations to compare the baselines with one another, considering all the scenarios discussed in Section~\ref{sec:experiments_settings}. As shown in Fig.~\ref{fig:20_nodes_obs}, NOTEARS and PC consistently outperform the other algorithms in terms of F1 score (we omit the results for the SHD due to lack of space, but the same trend holds). In order to avoid overcrowding the comparison with our approach, we will be using only these two algorithms as a benchmark for evaluation in Section~\ref{sec:result}.


\section{Results}\label{sec:result}

We now evaluate the structural accuracy of \textsc{DODO} on directed acyclic graphs. To ensure a fair comparison with observational baselines, both DODO and the baselines are subject to the same budget. Note that we have fixed the budget levels independently of the number of nodes.
This choice implies that, when all nodes are eligible for intervention, the per-node interventional budget available to \textsc{DODO} is effectively larger in smaller graphs. Larger budgets reduce variance in both phases, strengthen the t-test used to flag putative causal effects, and stabilize the conditional checks used to prune indirect paths.

\begin{figure}[!t]
  \centering
  \includegraphics[width=1\linewidth]{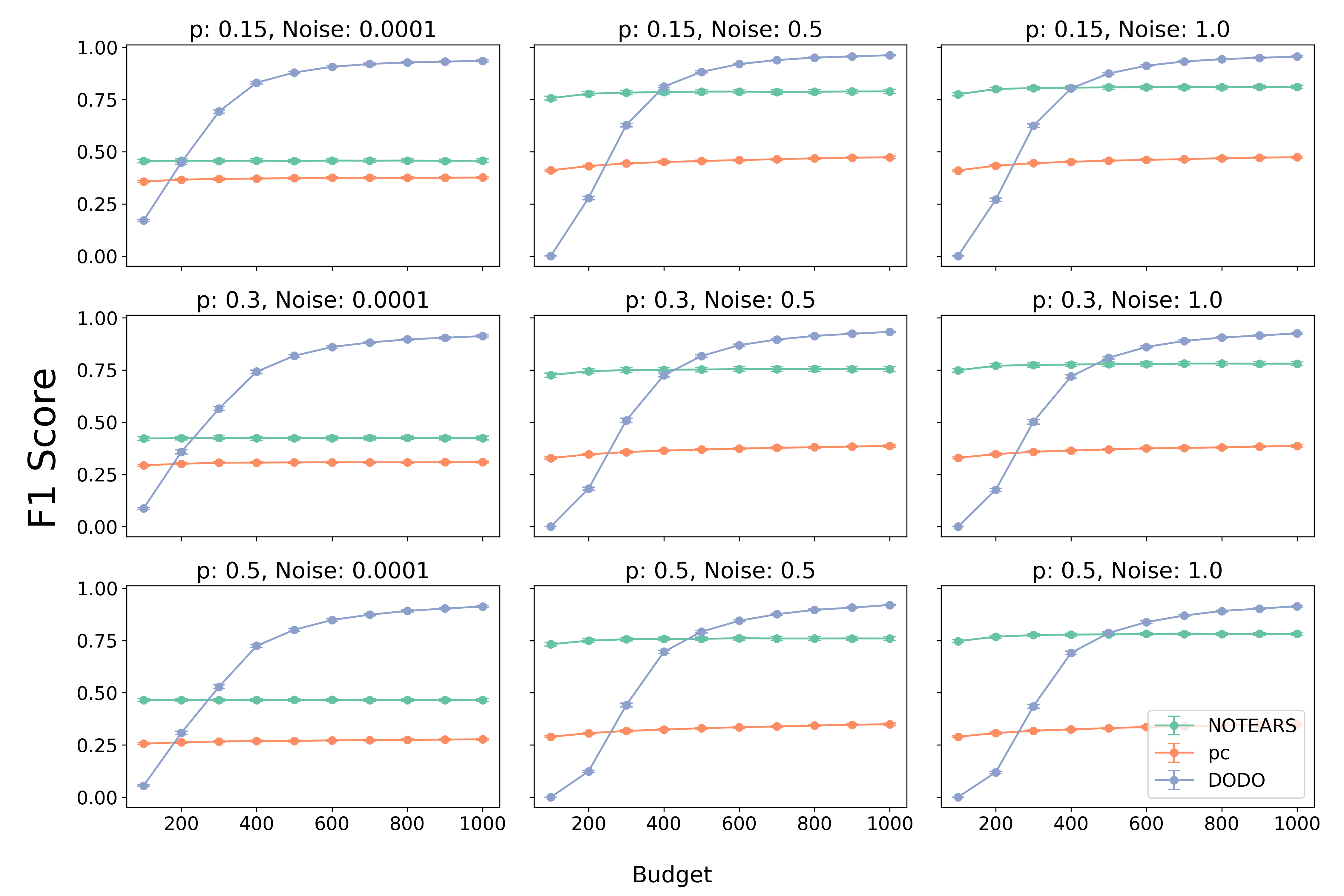}
  \caption{F\textsubscript{1} score for 20-node graphs; mean \(\pm\) 95\% confidence interval.}
  \label{fig:20_nodes_overview_line_f1}
\end{figure}

Fig.~\ref{fig:20_nodes_overview_line_f1} reports the F1 scores of DODO and the baseline methods for 20-node DAGs generated as described in Section~\ref{sec:dag_generation}.
Under this configuration, \textsc{DODO} exhibits a consistently rising F1 trajectory as the budget grows and it surpasses purely observational baselines while the associated uncertainty bands narrow in step with the increased replication. The curves for observational baselines show only a modest improvement as the budget increases, and this improvement remains limited. The underlying reason lies in the intrinsic limitations of observational approaches to causal discovery. While larger budgets enhance statistical discriminative power, certain relationships can only be uncovered through interventions~\cite{Pearl_2009}. Observing the figure from top to bottom, we see that DODO does not suffer from increasing edge density (larger~$p$), and the same holds for NOTEARS. In contrast, PC performs slightly better when the graph is sparser (as sparser graphs contain fewer statistically challenging causal dependencies).
Looking from left to right, we can observe the effects of increasing noise. 
The algorithms react differently to noise: PC shows only a slight improvement for higher noise levels, while NOTEARS experiences a drastic performance gain. This suggests that noise becomes a characteristic of the variable itself, making its overall effects easier to detect.
DODO is not particularly sensitive to noise and maintains consistently strong performance across all noise levels, except when the budget for interventions is too low.

\begin{figure}[!t]
  \centering
  \includegraphics[width=1\linewidth]{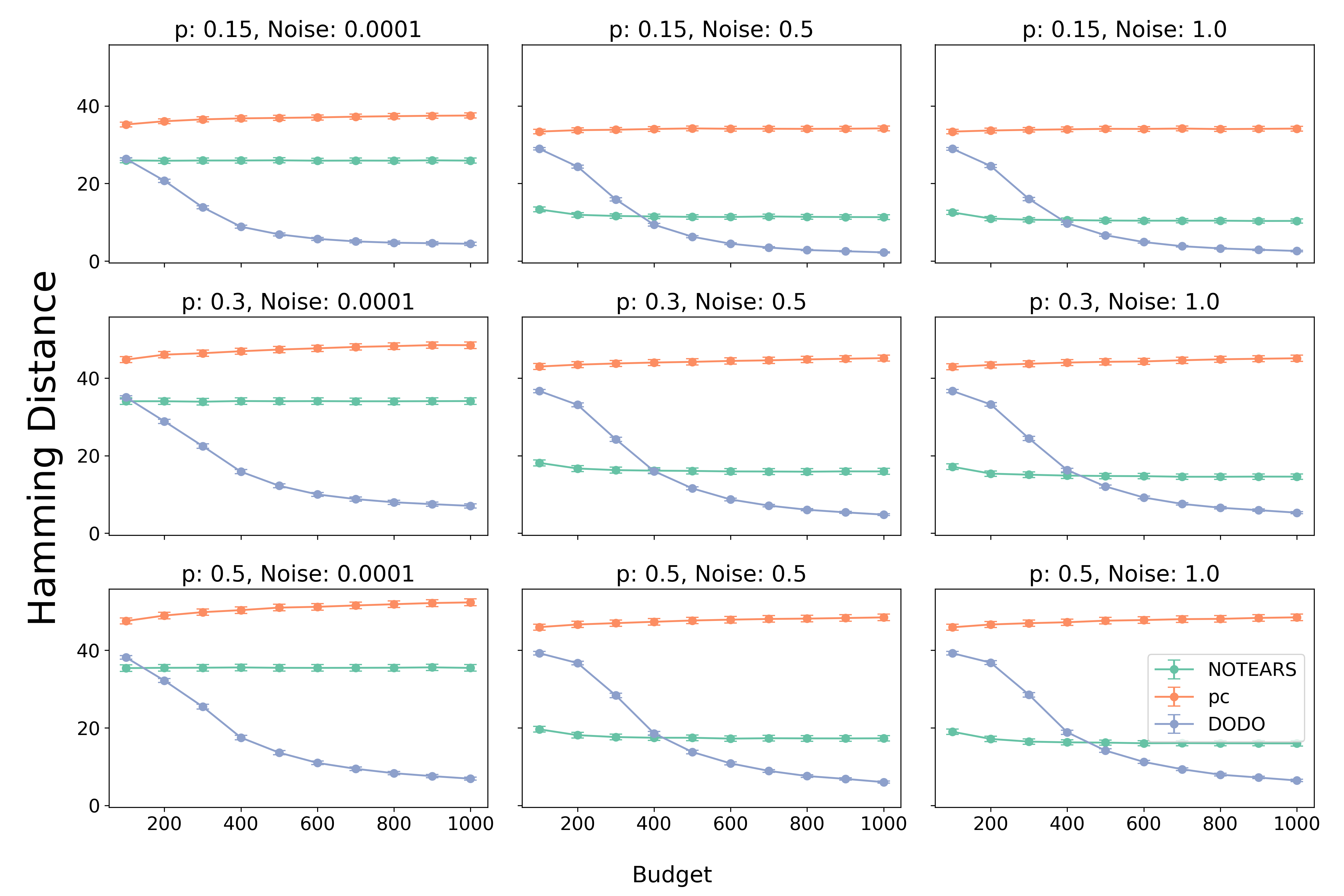}
  \caption{Structural Hamming Distance for 20-node graphs; mean \(\pm\) 95\% confidence interval.}
  \label{fig:20_nodes_overview_line_Hamming_distance}
\end{figure}

In the same twenty-node setting, the Structural Hamming Distance (Fig.~\ref{fig:20_nodes_overview_line_Hamming_distance}) decreases as budget increases, reflecting fewer spurious edges and more effective removal of transitive configurations such as two-step chains that can masquerade as direct links. The improvement appears once the per-node interventional replication (i.e., the number of samples the Agent can obtain for each intervention, under the given budget constraint) exceeds a threshold and then continues smoothly toward a plateau, indicating that the algorithm has acquired sufficient evidence to separate direct influences from mediated ones. When the budget is very small, the per-node observational and interventional replication becomes the limiting factor: the detection stage can still succeed for large effects, but the pruning stage, which conditions on sets that grow with in-degree in denser graphs, lacks the sample support required to reject indirect paths reliably, so false positives accumulate, precision drops, and the Structural Hamming Distance rises. These behaviors are visible in Fig.~\ref{fig:20_nodes_overview_line_Hamming_distance}, where the SHD values are uniformly lower for \textsc{DODO} once replication is sufficient, and the confidence bands tighten as variability across random topologies and noise realizations is averaged out by repeated epochs. The trends observed in the previous figure about the effect of noise and edge density are confirmed here.

\begin{figure}[!t]
  \centering
  \includegraphics[width=1\linewidth]{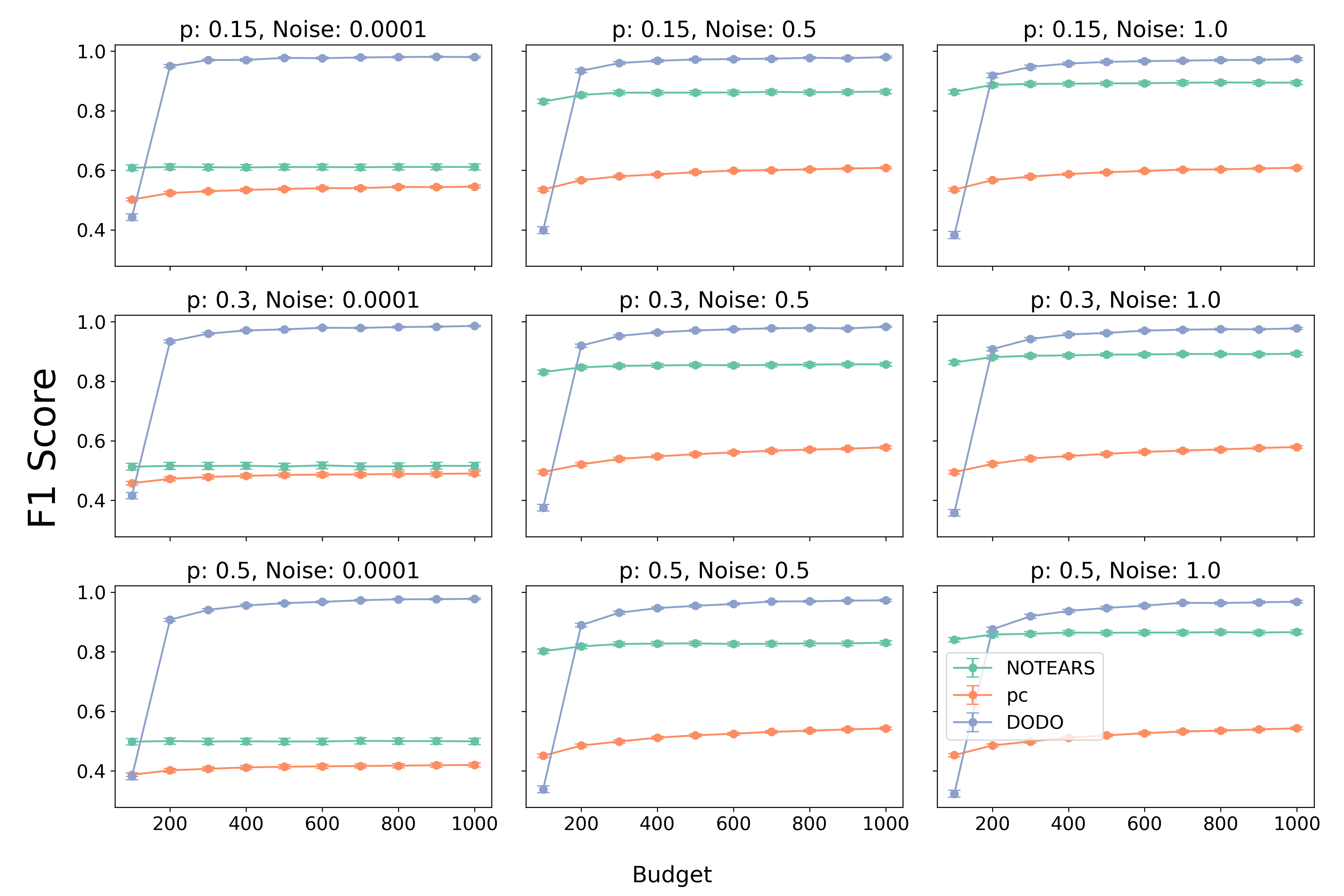}
  \caption{F\textsubscript{1} score for 10-node graphs; mean \(\pm\) 95\% confidence interval.}
  \label{fig:10_nodes_overview_line_f1}
\end{figure}

\begin{figure}[!t]
  \centering
  \includegraphics[width=1\linewidth]{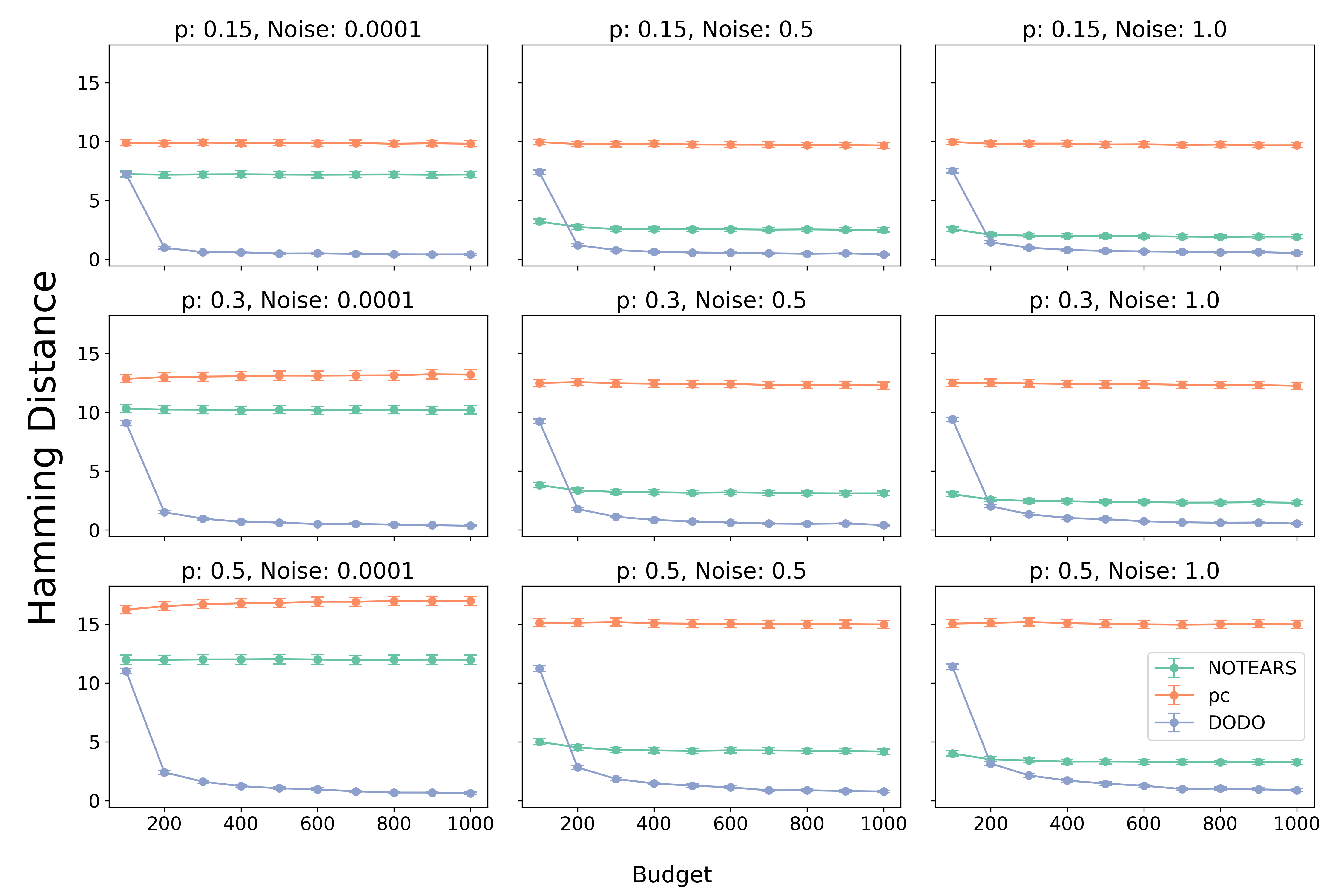}
  \caption{Structural Hamming Distance for 10-node graphs; mean \(\pm\) 95\% confidence interval.}
  \label{fig:10_nodes_overview_line_Hamming_distance}
\end{figure}

Based on our discussion so far, we expect smaller causal graphs to be easier to predict under the same budget constraints. This expectation is confirmed in Fig.~\ref{fig:10_nodes_overview_line_f1}–\ref{fig:10_nodes_overview_line_Hamming_distance}, which reports the results for 10-node graphs.
As a consequence of maintaining the total budget independent of the number of nodes, DODO gathers more post-intervention samples per targeted node, runs t-tests on larger samples, and enters the pruning stage with more stable conditional estimates. The result is a clear rise in the F1 score across budgets, accompanied by visibly narrower uncertainty bands that indicate stabilized decisions across random initializations and graph topologies. Fig.~\ref{fig:10_nodes_overview_line_f1} illustrates these gains and shows that performance improves smoothly as replication increases, mirroring the twenty-node case but at consistently higher levels because each node receives a larger share of the interventional effort.
The same mechanism drives the Structural Hamming Distance-based results: with more post-intervention samples per node, the pruning stage rejects indirect paths more decisively, and the learned graph aligns more closely with the ground truth. Fig.~\ref{fig:10_nodes_overview_line_Hamming_distance} reports the corresponding Structural Hamming Distances for ten-node graphs and shows uniformly lower values at comparable budgets relative to the twenty-node case, confirming that additional per node replication translates directly into superior topological fidelity. In budget-constrained regimes, a simple mitigation might be restricting the active intervention set to a subset of nodes so that each selected node receives more interventional replication; this reallocation would leave the observational phase untouched and would raise the power of both detection and pruning, but at the cost of exploring a subset of the nodes of the graph.

In summary, DODO displays superior causal discovery performance in well-resourced regimes across both metrics, but exhibits performance degradation when the intervention budget is insufficient to support robust pruning. This motivates adaptive strategies that adjust the number of interventions dynamically,  e.g., concentrating the reallocation of available samples to a smaller subset of nodes, thus maintaining statistical power and preserving identifiability, albeit for a limited part of the overall graph.

\section{Conclusion}\label{sec:conclusion}

In this work, we have presented DODO, a novel algorithmic framework designed for an autonomous Agent to infer the underlying causal structure of its environment, represented as a Directed Acyclic Graph (DAG) $\mathcal{G}_{DAG}$. The methodology is structured on a sequence of interventions, wherein the Agent systematically manipulates individual system variables. The resultant effects are quantified via statistical hypothesis testing, specifically two-sample t-tests, to establish an initial set of candidate causal links, $\mathcal{E}_{\text{cand}}$. The second step of DODO is its subsequent pruning of indirect causal connection phase, which employs partial correlation analysis to disambiguate direct causal influences from indirect, mediated effects, thereby refining the inferred graph structure to $\mathcal{G}_{\text{Agent}}$. Our empirical validation was conducted within a simulated environment governed by a linear Structural Equation Model with additive Gaussian noise, defined over DAGs of varying sizes and densities. The extensive simulation results demonstrate that, provided an adequate budget of interaction epochs ($K$), DODO consistently outperforms established observational baselines, including the Peter-Clark (PC) algorithm and NOTEARS. This superiority is quantitatively established through significantly higher F$_1$ scores and lower Structural Hamming Distances. 
Furthermore, these results underscore the intrinsic advantage of leveraging interventional data to resolve causal ambiguities that are intractable from observational data. Therefore, DODO establishes a robust foundation for interventional causal discovery and serves as a promising step towards developing more sophisticated, resource-aware autonomous agents capable of constructing accurate causal models of complex systems.

\section*{Acknowledgment}

This work was partially supported by SoBigData.it. SoBigData.it receives funding from European Union – NextGenerationEU – National Recovery and Resilience Plan (Piano Nazionale di Ripresa e Resilienza, PNRR) – Project: “SoBigData.it – Strengthening the Italian RI for Social Mining and Big Data Analytics” – Prot. IR0000013 – Avviso n. 3264 del 28/12/2021.
The work of M. Gregorini and L. Valerio is partly supported by PNRR - M4C2 - Investimento 1.3, Partenariato Esteso PE00000013 - ``FAIR - Future Artificial Intelligence Research'' - Spoke 1 "Human-centered AI", funded by the European Commission under the NextGeneration EU programme.
C. Boldrini was also supported by PNRR - M4C2 - Investimento 1.4, Centro Nazionale CN00000013 - "ICSC - National Centre for HPC, Big Data and Quantum Computing" - Spoke 6, funded by the European Commission under the NextGeneration EU programme.

\bibliographystyle{vancouver}
\bibliography{bibliography_phd_Gregorini}

\begin{thebibliography}{10}

\bibitem{LeCun_2015}
LeCun Y, Bengio Y, Hinton G.
\newblock Deep Learning.
\newblock Nature. 2015;521(7553):436-44.
\newblock Available from: \url{https://www.nature.com/articles/nature14539}.

\bibitem{Rudin_2019}
Rudin C.
\newblock Stop Explaining Black Box Machine Learning Models for High-Stakes Decisions and Use Interpretable Models Instead.
\newblock Nature Machine Intelligence. 2019;1(5):206-15.
\newblock Available from: \url{https://doi.org/10.1038/s42256-019-0048-x}.

\bibitem{YoshuaBengioRevered}
Yoshua Bengio, Revered Architect of {AI}, Has Some Ideas About What to Build Next - {IEEE} Spectrum;.

\bibitem{Pearl_2009}
Pearl J.
\newblock Causality: Models, Reasoning and Inference.
\newblock 2nd ed. Cambridge University Press; 2009.
\newblock Available from: \url{https://bayes.cs.ucla.edu/BOOK-2K/}.

\bibitem{hernan2023causal}
Hernan MA, Robins JM.
\newblock Causal Inference: What If.
\newblock Chapman \& Hall/CRC Monographs on Statistics \& Applied Probab. CRC Press; 2023.

\bibitem{Guo_2020}
Guo R, Cheng L, Li J, Hahn PR, Liu H.
\newblock A Survey of Learning Causality with Data: Problems and Methods.
\newblock ACM Computing Surveys. 2020 Jul;53(4):1–37.

\bibitem{zeng2023survey}
Zeng Y, Cai R, Sun F, Huang L, Hao Z. A Survey on Causal Reinforcement Learning; 2023.

\bibitem{deng2023causal}
Deng Z, Jiang J, Long G, Zhang C. Causal Reinforcement Learning: A Survey; 2023.

\bibitem{scholkopf2022causality}
Sch{\"o}lkopf B.
\newblock Causality for machine learning.
\newblock In: Probabilistic and Causal Inference: The Works of Judea Pearl; 2022. p. 765-804.

\bibitem{zhu2020causal}
Zhu S, Ng I, Chen Z. Causal Discovery with Reinforcement Learning; 2020.

\bibitem{dasgupta2019causal}
Dasgupta I, Wang J, Chiappa S, Mitrovic J, Ortega P, Raposo D, et~al.. Causal Reasoning from Meta-reinforcement Learning; 2019.

\bibitem{Spirtes_2000}
Spirtes P, Glymour C, Scheines R.
\newblock Causation, Prediction, and Search.
\newblock MIT Press; 2000.
\newblock Available from: \url{https://link.springer.com/book/10.1007/978-1-4612-2748-9}.

\bibitem{chauhan2025a}
Chauhan V, Dhami D, Gao B, Wang X, Clifton L, Clifton D. Beyond correlations: the necessity and the challenges of causal AI. techRxiv; 2025.

\bibitem{10.1145/3665494}
Rawal A, Raglin A, Rawat DB, Sadler BM, McCoy J.
\newblock Causality for Trustworthy Artificial Intelligence: Status, Challenges and Perspectives.
\newblock ACM Comput Surv. 2025 Feb;57(6).
\newblock Available from: \url{https://doi.org/10.1145/3665494}.

\bibitem{genomic}
König H, Frank D, Baumann M, Heil R.
\newblock AI models and the future of genomic research and medicine: True sons of knowledge?
\newblock BioEssays. 2021;43(10):2100025.
\newblock Available from: \url{https://onlinelibrary.wiley.com/doi/abs/10.1002/bies.202100025}.

\bibitem{causality_engineering}
Naser MZ.
\newblock Causality and causal inference for engineers: Beyond correlation, regression, prediction and artificial intelligence.
\newblock WIREs Data Mining and Knowledge Discovery. 2024;14(4):e1533.
\newblock Available from: \url{https://wires.onlinelibrary.wiley.com/doi/abs/10.1002/widm.1533}.

\bibitem{Shanmugam2015SmallInt}
Shanmugam K, Kocaoglu M, Dimakis AG, Vishwanath S.
\newblock Learning Causal Graphs with Small Interventions.
\newblock In: Advances in Neural Information Processing Systems 28 (NeurIPS 2015); 2015. p. 3195-203.
\newblock Available from: \url{https://papers.nips.cc/paper/5909-learning-causal-graphs-with-small-interventions}.

\bibitem{Madumal_2020}
Madumal P, Miller T, Sonenberg L, Vetere F.
\newblock Explainable Reinforcement Learning Through a Causal Lens.
\newblock In: Proceedings of the AAAI Conference on Artificial Intelligence; 2020. p. 2493-500.
\newblock Available from: \url{https://doi.org/10.1609/aaai.v34i03.5631}.

\bibitem{zhu2022causaldyna}
Zhu D, Li LE, Elhoseiny M. CausalDyna: Improving Generalization of Dyna-style Reinforcement Learning via Counterfactual-Based Data Augmentation; 2022.

\bibitem{wang2023voyager}
Wang G, Xie Y, Jiang Y, Mandlekar A, Xiao C, Zhu Y, et~al.. Voyager: An Open-Ended Embodied Agent with Large Language Models; 2023.

\bibitem{applications_health}
Triantafyllidis~AK TA. Applications of Machine Learning in Real-Life Digital Health Interventions: Review of the Literature; 2023.

\bibitem{nie2023knowledge}
Nie W, Wen X, Liu J, Chen J, Wu J, Jin G, et~al.
\newblock Knowledge-Enhanced Causal Reinforcement Learning Model for Interactive Recommendation.
\newblock IEEE Transactions on Multimedia. 2023.

\bibitem{pmlr-v119-zhang20a}
Zhang J.
\newblock Designing Optimal Dynamic Treatment Regimes: A Causal Reinforcement Learning Approach.
\newblock In: III HD, Singh A, editors. Proceedings of the 37th International Conference on Machine Learning. vol. 119 of Proceedings of Machine Learning Research. PMLR; 2020. p. 11012-22.

\bibitem{mental_health_interventions}
Morrison LG, Hargood C, Pejovic V, Geraghty AWA, Lloyd S, Goodman N, et~al.
\newblock The Effect of Timing and Frequency of Push Notifications on Usage of a Smartphone-Based Stress Management Intervention: An Exploratory Trial.
\newblock PLOS ONE. 2017 01;12(1):1-15.

\bibitem{SpirtesGlymour1991PC}
Spirtes P, Glymour C.
\newblock An Algorithm for Fast Recovery of Sparse Causal Graphs.
\newblock Social Science Computer Review. 1991;9(1):62-72.

\bibitem{spirtes2001anytime}
Spirtes P.
\newblock An Anytime Algorithm for Causal Inference.
\newblock In: Richardson TS, Jaakkola TS, editors. Proceedings of the Eighth International Workshop on Artificial Intelligence and Statistics. vol.~R3 of Proceedings of Machine Learning Research. PMLR; 2001. p. 278-85.

\bibitem{Colombo2012RFCI}
Colombo D, Maathuis MH, Kalisch M, Richardson TS.
\newblock Learning High-Dimensional Directed Acyclic Graphs with Latent and Selection Variables.
\newblock The Annals of Statistics. 2012;40(1):294-321.

\bibitem{Zhang2008FCI}
Zhang J.
\newblock On the Completeness of Orientation Rules for Causal Discovery in the Presence of Latent Confounders and Selection Bias.
\newblock Artificial Intelligence. 2008;172(16--17):1873-96.
\newblock Available from: \url{https://www.sciencedirect.com/science/article/pii/S0004370208001008}.

\bibitem{Chickering_2002}
Chickering DM.
\newblock Optimal Structure Identification with Greedy Search.
\newblock Journal of Machine Learning Research. 2002;3:507-54.
\newblock Available from: \url{https://www.jmlr.org/papers/v3/chickering02a.html}.

\bibitem{Ramsey2017FGES}
Ramsey JD, Glymour M, Sanchez-Romero R, Glymour C.
\newblock A Million Variables and More: The Fast Greedy Equivalence Search Algorithm for Learning High-Dimensional Graphical Causal Models.
\newblock International Journal of Data Science and Analytics. 2017;3(2):121-9.
\newblock Available from: \url{https://link.springer.com/content/pdf/10.1007/s41060-016-0032-z.pdf}.

\bibitem{Zheng2018NOTEARS}
Zheng X, Aragam B, Ravikumar P, Xing EP.
\newblock DAGs with {NO TEARS}: Continuous Optimization for Structure Learning.
\newblock In: Advances in Neural Information Processing Systems 31 (NeurIPS 2018); 2018. .

\bibitem{Shimizu2006LiNGAM}
Shimizu S, Hoyer PO, Hyv{\"a}rinen A, Kerminen A.
\newblock A Linear Non-Gaussian Acyclic Model for Causal Discovery.
\newblock Journal of Machine Learning Research. 2006;7:2003-30.
\newblock Available from: \url{https://jmlr.org/papers/volume7/shimizu06a/shimizu06a.pdf}.

\bibitem{Hoyer2009ANM}
Hoyer PO, Janzing D, Mooij JM, Peters J, Sch{\"o}lkopf B.
\newblock Nonlinear Causal Discovery with Additive Noise Models.
\newblock In: Advances in Neural Information Processing Systems 21 (NeurIPS 2008); 2009. p. 689-96.
\newblock Available from: \url{https://papers.neurips.cc/paper/2008/file/f7664060cc52bc6f3d620bcedc94a4b6-Paper.pdf}.

\bibitem{Peters2014ANM}
Peters J, Mooij JM, Janzing D, Sch{\"o}lkopf B.
\newblock Causal Discovery with Continuous Additive Noise Models.
\newblock Journal of Machine Learning Research. 2014;15:2009-53.
\newblock Available from: \url{https://jmlr.org/papers/volume15/peters14a/peters14a.pdf}.

\bibitem{Zhang2011KCI}
Zhang K, Peters J, Janzing D, Sch{\"o}lkopf B.
\newblock Kernel-based Conditional Independence Test and Application in Causal Discovery.
\newblock In: Proceedings of the 27th Conference on Uncertainty in Artificial Intelligence (UAI 2011); 2011. p. 804-13.
\newblock Available from: \url{https://is.mpg.de/publications/zhangpjs2011}.

\bibitem{Gretton2005HSIC}
Gretton A, Herbrich R, Smola A, Bousquet O, Sch{\"o}lkopf B.
\newblock Measuring Statistical Dependence with Hilbert--Schmidt Norms.
\newblock Journal of Machine Learning Research. 2005;6:2075-129.
\newblock Available from: \url{https://www.gatsby.ucl.ac.uk/~gretton/papers/GreBouSmoSch05.pdf}.

\bibitem{Gretton2007HSIC}
Gretton A, Fukumizu K, Teo CH, Song L, Sch{\"o}lkopf B, Smola AJ.
\newblock A Kernel Statistical Test of Independence.
\newblock In: Advances in Neural Information Processing Systems 20 (NeurIPS 2007); 2007. p. 585-92.
\newblock Available from: \url{https://proceedings.neurips.cc/paper/2007/file/d5cfead94f5350c12c322b5b664544c1-Paper.pdf}.

\bibitem{Hauser2012GIES}
Hauser A, B{\"u}hlmann P.
\newblock Characterization and Greedy Learning of Interventional Markov Equivalence Classes of {DAG}s.
\newblock Journal of Machine Learning Research. 2012;13:2409-64.
\newblock Available from: \url{https://jmlr.org/papers/volume13/hauser12a/hauser12a.pdf}.

\bibitem{Eberhardt2006Nminus1}
Eberhardt F, Glymour C, Scheines R.
\newblock N-1 Experiments Suffice to Determine the Causal Relations Among N Variables.
\newblock In: Innovations in Machine Learning. vol. 194 of Studies in Fuzziness and Soft Computing. Berlin, Heidelberg: Springer; 2006. p. 97-112.
\newblock Available from: \url{https://link.springer.com/chapter/10.1007/3-540-33486-6_4}.

\bibitem{Eberhardt2012LogN}
Eberhardt F, Glymour C, Scheines R.
\newblock On the Number of Experiments Sufficient and in the Worst Case Necessary to Identify All Causal Relations Among N Variables.
\newblock In: Proceedings of the Twenty-First Conference on Uncertainty in Artificial Intelligence (UAI 2005); 2005. Preprint arXiv:1207.1389.
\newblock Available from: \url{https://arxiv.org/abs/1207.1389}.

\bibitem{HeGeng2008JMLR}
He Y, Geng Z.
\newblock Active Learning of Causal Networks with Intervention Experiments and Optimal Designs.
\newblock Journal of Machine Learning Research. 2008;9:2523-47.
\newblock Available from: \url{https://jmlr.csail.mit.edu/papers/volume9/he08a/he08a.pdf}.

\bibitem{Hyttinen2013JMLR}
Hyttinen A, Eberhardt F, Hoyer PO.
\newblock Experiment Selection for Causal Discovery.
\newblock Journal of Machine Learning Research. 2013;14:3041-71.
\newblock Available from: \url{https://www.jmlr.org/papers/volume14/hyttinen13a/hyttinen13a.pdf}.

\bibitem{Hauser2014IJAR}
Hauser A, B{\"u}hlmann P.
\newblock Two Optimal Strategies for Active Learning of Causal Models from Interventions.
\newblock International Journal of Approximate Reasoning. 2014;55(4):926-39.
\newblock Available from: \url{https://www.sciencedirect.com/science/article/pii/S0888613X13002879}.

\bibitem{Mooij2020JCI}
Mooij JM, Magliacane S, Claassen T.
\newblock Joint Causal Inference from Multiple Contexts.
\newblock Journal of Machine Learning Research. 2020;21(99):1-108.
\newblock Available from: \url{https://jmlr.org/papers/v21/17-123.html}.

\bibitem{Brouillard2020DCDI}
Brouillard P, Lachapelle S, Lacoste A, Lacoste-Julien S, Drouin A.
\newblock Differentiable Causal Discovery from Interventional Data.
\newblock In: Advances in Neural Information Processing Systems 33 (NeurIPS 2020); 2020. Available from: \url{https://proceedings.neurips.cc/paper/2020/file/f8b7aa3a0d349d9562b424160ad18612-Paper.pdf}.

\bibitem{Jaber2020Soft}
Jaber A, Kocaoglu M, Shanmugam K, Bareinboim E.
\newblock Causal Discovery from Soft Interventions with Unknown Targets: Characterization and Learning.
\newblock In: Advances in Neural Information Processing Systems 33 (NeurIPS 2020); 2020. Available from: \url{https://proceedings.neurips.cc/paper/2020/hash/6cd9313ed34ef58bad3fdd504355e72c-Abstract.html}.

\bibitem{Lattimore2016CausalBandits}
Lattimore F, Lattimore T, Reid MD.
\newblock Causal Bandits: Learning Good Interventions via Causal Inference.
\newblock In: Advances in Neural Information Processing Systems 29 (NeurIPS 2016); 2016. Available from: \url{https://proceedings.neurips.cc/paper_files/paper/2016/file/b4288d9c0ec0a1841b3b3728321e7088-Paper.pdf}.

\bibitem{Scherrer2021ActiveInt}
Scherrer N, Bilaniuk O, Annadani Y, Goyal A, Schwab P, Sch{\"o}lkopf B, et~al.. Learning Neural Causal Models with Active Interventions; 2021.
\newblock Available from: \url{https://arxiv.org/pdf/2109.02429}.

\bibitem{Glymour2019Review}
Glymour C, Zhang K, Spirtes P.
\newblock Review of Causal Discovery Methods Based on Graphical Models.
\newblock Frontiers in Genetics. 2019;10:524.

\bibitem{Sachs2005Science}
Sachs K, Perez O, Pe'er D, Lauffenburger DA, Nolan GP.
\newblock Causal Protein-Signaling Networks Derived from Multiparameter Single-Cell Data.
\newblock Science. 2005;308(5721):523-9.

\bibitem{LauritzenSpiegelhalter1988}
Lauritzen SL, Spiegelhalter DJ.
\newblock Local Computation with Probabilities on Graphical Structures and their Application to Expert Systems.
\newblock Journal of the Royal Statistical Society: Series B (Methodological). 1988;50(2):157-224.
\newblock Asia network (8 variables) widely reused in BN benchmarks.

\bibitem{Cooper}
Cooper GF, Yoo C.
\newblock Causal discovery from a mixture of experimental and observational data.
\newblock In: Proceedings of the Fifteenth Conference on Uncertainty in Artificial Intelligence. UAI'99. San Francisco, CA, USA: Morgan Kaufmann Publishers Inc.; 1999. p. 116–125.

\bibitem{Bansal_2007}
Bansal M, Belcastro V, Ambesi-Impiombato A, di~Bernardo D.
\newblock How to infer gene networks from expression profiles.
\newblock Molecular Systems Biology. 2007;3:78.
\newblock Available from: \url{https://doi.org/10.1038/msb4100120}.

\bibitem{Aicher_2014}
Aicher C, Jacobs AZ, Clauset A.
\newblock Learning Latent Block Structure in Weighted Networks.
\newblock Journal of Complex Networks. 2014;3(2):221-48.
\newblock Available from: \url{https://doi.org/10.1093/comnet/cnu026}.

\bibitem{Robinson_1973}
Robinson RW.
\newblock Counting labeled acyclic digraphs.
\newblock New Directions in the Theory of Graphs. 1973:239-73.
\newblock Available from: \url{https://doi.org/10.1016/B978-0-12-543750-4.50018-5}.

\bibitem{Koller_Friedman_2009}
Koller D, Friedman N.
\newblock Probabilistic Graphical Models: Principles and Techniques.
\newblock MIT Press; 2009.
\newblock Available from: \url{https://mitpress.mit.edu/books/probabilistic-graphical-models}.

\bibitem{Meka_Pitassi_2009}
Meka R, Pitassi T.
\newblock Randomized Decision Tree Complexity of Read-once Formulas.
\newblock In: Proceedings of the 24th Annual IEEE Conference on Computational Complexity; 2009. p. 193-204.
\newblock Available from: \url{https://doi.org/10.1109/CCC.2009.27}.

\bibitem{Dagum_Luby_1992}
Dagum P, Luby M.
\newblock Approximating Probabilistic Inference in Bayesian Belief Networks is NP-hard.
\newblock Artificial Intelligence. 1992;60(1):141-53.
\newblock Available from: \url{https://doi.org/10.1016/0004-3702(93)90036-B}.

\bibitem{PetersBuhlmann2014}
Peters J, B{\"u}hlmann P.
\newblock Identifiability of Gaussian structural equation models with equal error variances.
\newblock Biometrika. 2014;101(1):219-28.
\newblock Available from: \url{https://academic.oup.com/biomet/article/101/1/219/2364921}.

\bibitem{ParkKim2020}
Park G, Kim Y.
\newblock Identifiability of Gaussian linear structural equation models with homogeneous and heterogeneous error variances.
\newblock Journal of the Korean Statistical Society. 2020;49:276-92.
\newblock Available from: \url{https://link.springer.com/article/10.1007/s42952-019-00019-7}.

\bibitem{KalischBuhlmann2007}
Kalisch M, B{\"u}hlmann P.
\newblock Estimating High-Dimensional Directed Acyclic Graphs with the PC-Algorithm.
\newblock Journal of Machine Learning Research. 2007;8:613-36.
\newblock Available from: \url{https://jmlr.org/papers/volume8/kalisch07a/kalisch07a.pdf}.

\bibitem{GeigerHeckerman1994}
Geiger D, Heckerman D.
\newblock Learning Gaussian Networks.
\newblock In: Proceedings of the Tenth Conference on Uncertainty in Artificial Intelligence (UAI-94). San Francisco, CA: Morgan Kaufmann; 1994. p. 235-43.
\newblock Available from: \url{https://arxiv.org/abs/1302.6808}.

\bibitem{PetersJanzingSchoelkopf2017}
Peters J, Janzing D, Sch{\"o}lkopf B.
\newblock Elements of Causal Inference: Foundations and Learning Algorithms.
\newblock Cambridge, MA: The MIT Press; 2017.
\newblock Available from: \url{https://mitpress.mit.edu/9780262037310/elements-of-causal-inference/}.

\bibitem{lauritzen1996graphical}
Lauritzen SL.
\newblock Graphical Models.
\newblock Oxford: Clarendon Press, Oxford University Press; 1996.

\bibitem{carroll2006measurement}
Carroll RJ, Ruppert D, Stefanski LA, Crainiceanu CM.
\newblock Measurement Error in Nonlinear Models: A Modern Perspective.
\newblock 2nd ed. Boca Raton, FL: Chapman \& Hall/CRC; 2006.

\bibitem{feller1971introduction}
Feller W.
\newblock An Introduction to Probability Theory and Its Applications, Volume II.
\newblock 2nd ed. New York: John Wiley \& Sons; 1971.

\bibitem{bollen1989structural}
Bollen KA.
\newblock Structural Equations with Latent Variables.
\newblock New York: John Wiley \& Sons; 1989.

\bibitem{ParkMoonParkJeon2021}
Park G, Moon SJ, Park S, Jeon JJ.
\newblock Learning a High-dimensional Linear Structural Equation Model via $\ell_1$-Regularized Regression.
\newblock Journal of Machine Learning Research. 2021;22(102):1-41.
\newblock Available from: \url{https://jmlr.org/papers/v22/20-1005.html}.

\bibitem{Mooij2016Benchmark}
Mooij JM, Peters J, Janzing D, Zscheischler J, Sch{\"o}lkopf B.
\newblock Distinguishing Cause from Effect Using Observational Data: Methods and Benchmarks.
\newblock Journal of Machine Learning Research. 2016;17(32):1-102.

\bibitem{barabasi2013network}
Barab{\'a}si AL.
\newblock Network science.
\newblock Philosophical Transactions of the Royal Society A: Mathematical, Physical and Engineering Sciences. 2013;371(1987):20120375.

\bibitem{Granger_1969}
Granger CWJ.
\newblock Investigating Causal Relations by Econometric Models and Cross-Spectral Methods.
\newblock Econometrica. 1969;37(3):424-38.
\newblock Available from: \url{https://www.jstor.org/stable/1912791}.

\end{thebibliography}

\end{document}